\documentclass[letterpaper, 10 pt, conference]{ieeeconf}  

\IEEEoverridecommandlockouts                              

\usepackage{graphicx}
\usepackage{color}
\usepackage{colortbl}
\usepackage{multirow}
\usepackage{pifont}
\usepackage{amsmath}
\usepackage{amssymb}
\usepackage{siunitx}
\usepackage[]{footmisc}

\title{\LARGE \bf
Assessing YOLACT++ for real time and robust instance segmentation \\ of medical instruments in endoscopic procedures}

\author{Juan Carlos Ángeles Cerón$^{1}$, Leonardo Chang$^{1}$, Gilberto Ochoa Ruiz$^{1}$, Sharib Ali$^{2}$

\thanks{$^{1}$Tecnológico de Monterrey, School of Engineering and Sciences, Av. Eugenio Garza Sada 2501, Monterrey, N.L., México 64849, Mexico}%
\thanks{$^{2}$ Institute of Biomedical Engineering and Big Data Institute, Old Road Campus, University of Oxford, Oxford, UK}%
\thanks{C. Authors:jcarlos.angelesc@gmail.com, gilberto.ochoa@tec.mx }%

\thanks{\textbf{*Preprint  under  review  for  EMBC’21  following  IEEE  guidelines}}
}

\begin{document}

\maketitle
\thispagestyle{empty}
\pagestyle{empty}

\begin{abstract}
Image-based tracking of laparoscopic instruments plays a fundamental role in computer and robotic-assisted surgeries by aiding surgeons and increasing patient safety. Computer vision contests, such as the Robust Medical Instrument Segmentation (ROBUST-MIS) Challenge, seek to encourage the development of robust models for such purposes, providing large, diverse, and high-quality datasets. To date, most of the existing models for instance segmentation of medical instruments were based on two-stage detectors, which provide robust results but are nowhere near to the real-time, running at 5 frames-per-second (fps) at most. However, for the method to be clinically applicable, a real-time capability is utmost required along with high accuracy. In this paper, we propose the addition of attention mechanisms to the YOLACT architecture to allow real-time instance segmentation of instruments with improved accuracy on the ROBUST-MIS dataset. Our proposed approach achieves competitive performance compared to the winner of the 2019 ROBUST-MIS challenge in terms of robustness scores, obtaining 0.313 MI\_DSC and 0.338 MI\_NSD while reaching real-time performance at $\mathbf{>}$45 fps.
\end{abstract}

\indent \textbf{Keywords:} \textit {
Deep learning; laparoscopy; segmentation; surgical instrument, real-time.}

\section{Introduction}
Computer-assisted minimally invasive surgery such as endoscopy has grown in popularity over the past years.  However, due to the nature of these procedures, issues like limited field-of-view, extreme lighting conditions, lack of depth information, and difficulty in manipulating operating instruments demand strenuous amounts of effort from the surgeons \cite{Ross_2020}. Surgical data science applications could provide physicians with context-aware assistance during minimally invasive surgery to overcome these limitations and increase patient safety. One of the main forms of assistance is providing accurate tracking of medical instruments using instance segmentation methods. These systems are expected to be a crucial component in tasks ranging from surgical navigation, skill analysis, complication prediction, and other computer-integrated surgery (CIS) applications \cite{Maierhein_2021}. 

Nonetheless, instrument tracking methods are often deployed in difficult scenarios such as bleeding, over or underexposure, smoke, and reflections \cite{Bodenstedt_2018}. The net effect of these issues increases the missed detection rates in endoscopic surveillance, hampering the adoption of AI-based tools in this context \cite{ali2021}. Therefore, the development of robust techniques that can be effectively deployed in real endoscopy interventions is very much necessary. 

Endoscopic computer vision contests, such as the Robust Medical Instrument Segmentation (ROBUST-MIS) Challenge \cite{Ross_2020} represent important and necessary efforts to encourage the development of robust models for surgical instrument segmentation. They integrate the developments in computer-assisted surgeries and benchmark the generalization capabilities of the developed methods on different clinical scenarios. Furthermore, they provide large-high-quality datasets to overcome one of the main bottlenecks of developing robust methodologies, which is the lack of annotated data. 

Previous approaches for instance segmentation submitted to the 2019 ROBUST-MIS challenge, were exclusively based on two-stage detectors such as Mask R-CNN \cite{He_2018}. While these models exhibited good performances in terms of robustness, they all suffered from very high inference times averaging around 5 fps, preventing them from achieving real-time performances. Realistically, real-time performance is mandatory in order to fully exploit the capabilities of tracking applications in live surgeries.

In order to overcome these inference limitations while maintaining a robust performance in terms of tool segmentation results, we propose a new approach based on YOLACT++ \cite{Bolya_2020} equipped with attention modules on the multi-scale outputs of the CNN backbone and Feature Pyramid Network (FPN). The increased representation power achieved by using attention allows the extraction of more discriminant features while suppressing the less effective ones. 

\begin{figure}[t!]
    \centering
    \includegraphics[width=.5\textwidth]{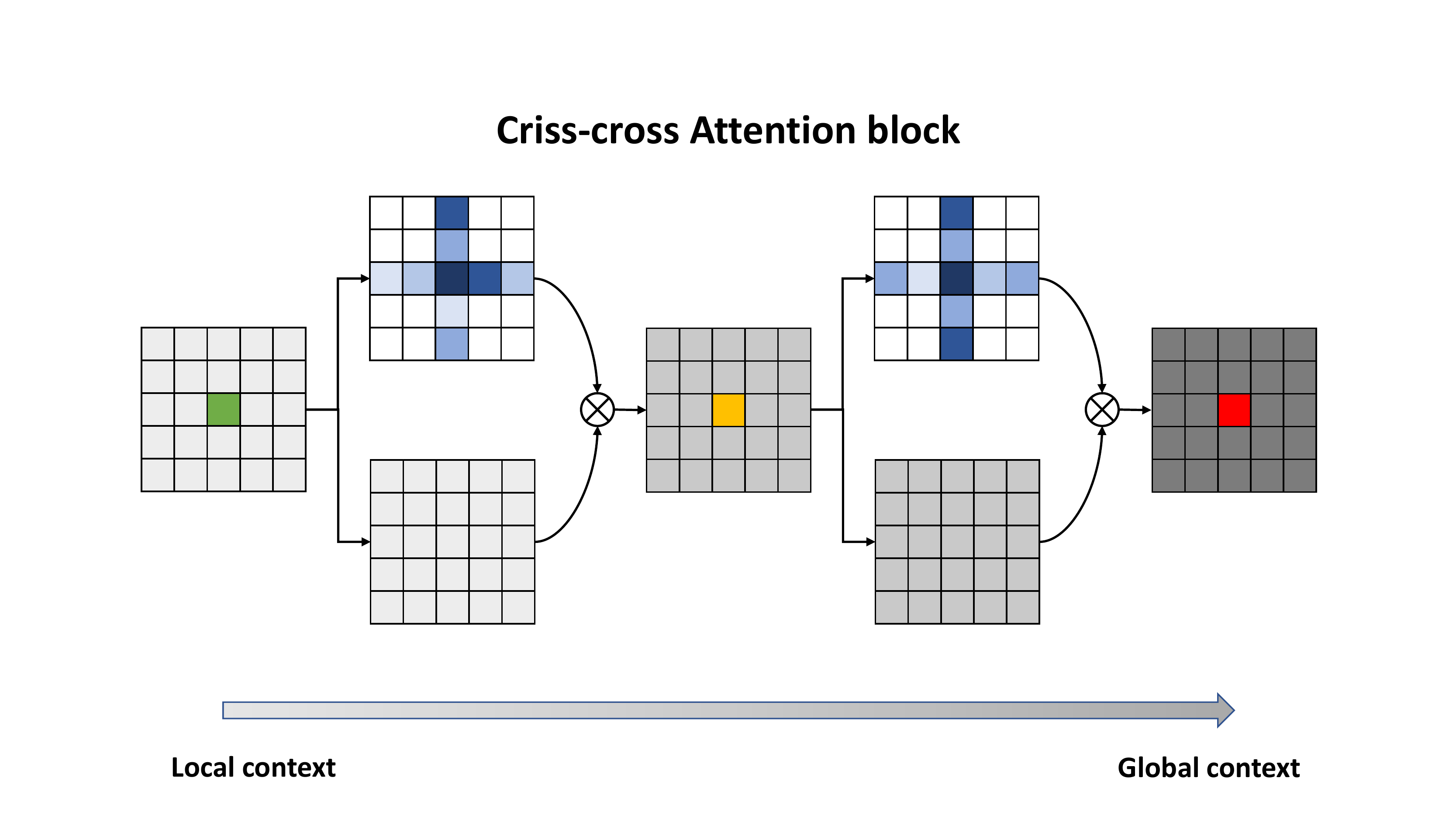}
    \caption{Diagram of the criss-cross attention module. For each position in the input feature map, the attention module generates a sparse attention map by aggregating information in the horizontal and vertical axes. After the second operation, each pixel has extracted context from all the pixels. }
    \label{fig:ccam}
\end{figure}

In this work, we evaluate the Criss-cross Attention Module (CCAM) \cite{Huang_2020}. CCAM, which is depicted in Figure \ref{fig:ccam}, recursively integrates global context across feature maps in a fast and clever criss-cross fashion. By integrating this attention mechanism, our proposed model outperforms previous approaches in the state-of-the-art by a slight margin, but it attains real-time performances, which makes our method clinically applicable both in inference time and robustness.

The rest of the paper is organized as follows. In Section \ref{sec:materials_&_methods} we discuss the ROBUST-MIS dataset, the context of the challenge, and the official metrics to assess robustness. Section \ref{sec:proposed_model} introduces our modifications to the YOLACT model to incorporate attention; also, we discuss our training and validation processes. In Section \ref{sec:results_&_discussion} we present our results and we discuss areas for further improvements. Finally, Section \ref{sec:conclusion} concludes the article.

\section{Materials and methods}
\label{sec:materials_&_methods}

\subsection{Materials}

The Heidelberg Colorectal Data Set \cite{Maierhein_2021} served as a basis for the ROBUST-MIS challenge. It comprises 30 surgical procedures from three different types of surgeries and includes detailed segmentation maps for the surgical instruments in more than 10,000 laparoscopic video frames. The generalization and performance of the submitted algorithms are typically assessed in three stages, with increasing levels of difficulty:

\begin{itemize}
	\item \textbf{Stage 1:} Test data is taken from the procedures from which the training data was extracted.
	\item \textbf{Stage 2:} Test data is taken from the same type of surgery as the training data but procedures not included in the training.
    \item \textbf{Stage 3:} Test data is taken from a different but similar type of surgery compared to the training data.
\end{itemize}

The detailed case distribution for each stage is presented in Table \ref{tab:data_distribution}. 

\begin{table}
\caption{Case distribution for each stage of the challenge \cite{Ross_2020}. 
}
\label{tab:data_distribution}
\begin{tabular}{lcccc}
\multicolumn{1}{c}{\textbf{Procedure}} &
  \textbf{Training} &
  \multicolumn{3}{c}{\textbf{Testing}} \\ \hline
 &
  \multicolumn{1}{l}{} &
  Stage 1 &
  Stage 2 &
  Stage 3 \\ \hline
Proctocolectomy &
  \begin{tabular}[c]{@{}c@{}}2,943 \end{tabular} &
  \begin{tabular}[c]{@{}c@{}}325 \end{tabular} &
  \begin{tabular}[c]{@{}c@{}}255 \end{tabular} &
  0 \\
Rectal resection &
  \begin{tabular}[c]{@{}c@{}}3,040 \end{tabular} &
  \begin{tabular}[c]{@{}c@{}}338 \end{tabular} &
  \begin{tabular}[c]{@{}c@{}}289 \end{tabular} &
  0 \\
Sigmoid resection* &
  0 &
  0 &
  0 &
  \begin{tabular}[c]{@{}c@{}}2,880 \end{tabular} \\ \hline
TOTAL &
  \begin{tabular}[c]{@{}c@{}}5,983 \end{tabular} &
  \begin{tabular}[c]{@{}c@{}}663 \end{tabular} &
  \begin{tabular}[c]{@{}c@{}}514 \end{tabular} &
  \begin{tabular}[c]{@{}c@{}}2,880 \end{tabular} 
\end{tabular}
\end{table}

\subsection{Metrics}

The two multi-instance segmentation metrics were used to assess the performance of the models. Multiple Instance Dice Similarity Coefficient (MI\_DSC) and Multiple Instance Normalized Surface Dice (MI\_NSD).
The DSC \cite{Dice_1945} is defined as the harmonic mean of precision and recall:

\begin{equation}
	DSC (Y, \hat{Y}) := \frac{2 \mid Y \cap \hat{Y} \mid}{\mid Y \mid + \mid \hat{Y} \mid},  
\end{equation}

Where $Y$ indicates the ground truth annotation and $\hat{Y}$ the corresponding prediction of an image frame.

Unlike DSC, which measures the overlap of volumes, the NSD measures the overlap of two mask borders \cite{ Nikolov_2021}. The metric uses a threshold that is related to the inter-rater variability of the annotators. According to \cite{Ross_2020}, their calculations resulted in a threshold of $\tau := 13$ for the challenge's data set. To calculate the MI\_DSC and MI\_NSD, matches of instrument instances were computed. Then, the resulting metric scores per instrument instance per image were aggregated by the mean.

Note that the challenge reports robustness and accuracy rankings. However, to compute accuracy, it is mandatory to know the per image results per participant, which are not available due to privacy issues. For this reason, we will be reporting only robustness rankings.

The robustness rankings pay particular attention in stage 3 of the challenge since it was built to test generalization, and focus on the worst-case performance of methods. For this reason, MI\_DSC and MI\_NSD are aggregated by the 5\% percentile instead of by the mean or median \cite{Ross_2020}.

\subsection{Data preprocessing}

A total of 996 frames with no visible instruments were removed from the training set, leaving 4,987 usable frames. An 85-15 percent split was made for training and validation purposes from this subset, respectively. 

Data augmentation techniques were heavily applied to introduce as much variability as possible and increase the model’s performance. The augmentation techniques used are random photometric distortions, random scaling, random sample cropping, and random mirroring.

\section{Proposed model}
\label{sec:proposed_model}

\subsection{Architecture of the proposed model}

\begin{figure}
    \centering
    \includegraphics[width=.5\textwidth]{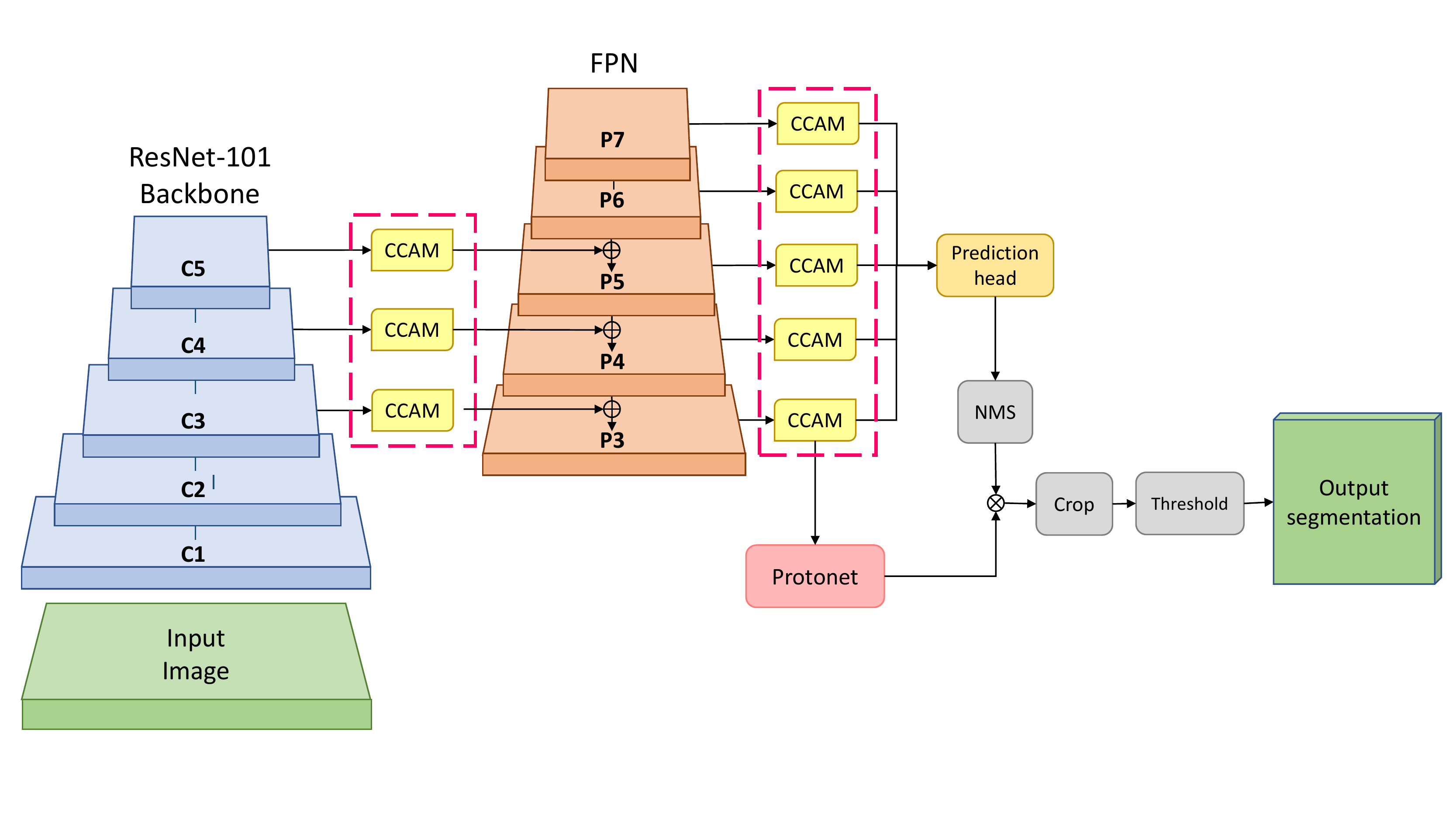}
    \caption{Proposed YOLACT++ architecture with criss-cross attention modules in ResNet-101 backbone + FPN. Note that certain modules are removed in some experiments. See Section \ref{sec:proposed_model} for further details.}
    \label{fig:architecture}
\end{figure}

In order to improve the robustness of the real-time YOLACT architecture used in our proposal, we introduce attention modules on the multi-scale outputs of the ResNet-101 backbone and the output features of the FPN (see Figure  \ref{fig:architecture}). Attention enables the network to focus on the most relevant features and avoid redundant use of information.

Our attention mechanism of choice was Criss-cross Attention Module (CCAM) \cite{Huang_2020}, specifically because of its fast, computationally efficient ($N\sqrt{N}$), and low GPU memory usage. These characteristics are crucial in order to introduce as little time-processing overhead as possible into the model and preserve real-time performance.

CCAM captures global contextual information in a similar fashion to the non-local module \cite{Wang_2018} by generating an attention map for each pixel in the input feature map and then transforming it into a refined feature map. However, instead of generating attention maps for each pixel-pair which is computationally expensive, for each pixel in a feature map, CCAM aggregates contextual information only in its horizontal and vertical directions, as shown in Figure \ref{fig:ccam}. By consecutively stacking two criss-cross attention modules, each pixel can collect contextual information from all pixels in a given feature map. Next, the contextually rich feature is concatenated and convolved with the original feature maps for feature fusion. Our experiments consisted of systematically adding the attention modules in two strategic locations in the network: first, in the backbone’s output features, next in the FPN output features, and ultimately on both locations. As a result, we came up with three models, \emph{CCAM-Backbone}, \emph{CCAM-FPN}, and \emph{CCAM-Full}, plus the attentionless baseline \emph{Base YOLACT++}.

\subsection{Training and validation}

We trained the models for up to 100,000 iterations with a learning rate of 0.001, momentum of 0.9, weight decay of \num{5e-4}, and batch size 16 in an NVIDIA DGX-1 system. The performance was assessed using the evaluation code for the challenge \cite{Ross_2019}. Moreover, the rankings were computed using the R package \emph{challengeR} \cite{Wiesenfarth_2019}.

\section{Results and discussion}
\label{sec:results_&_discussion}

\begin{figure}
    \centering
    \includegraphics[width=.5\textwidth]{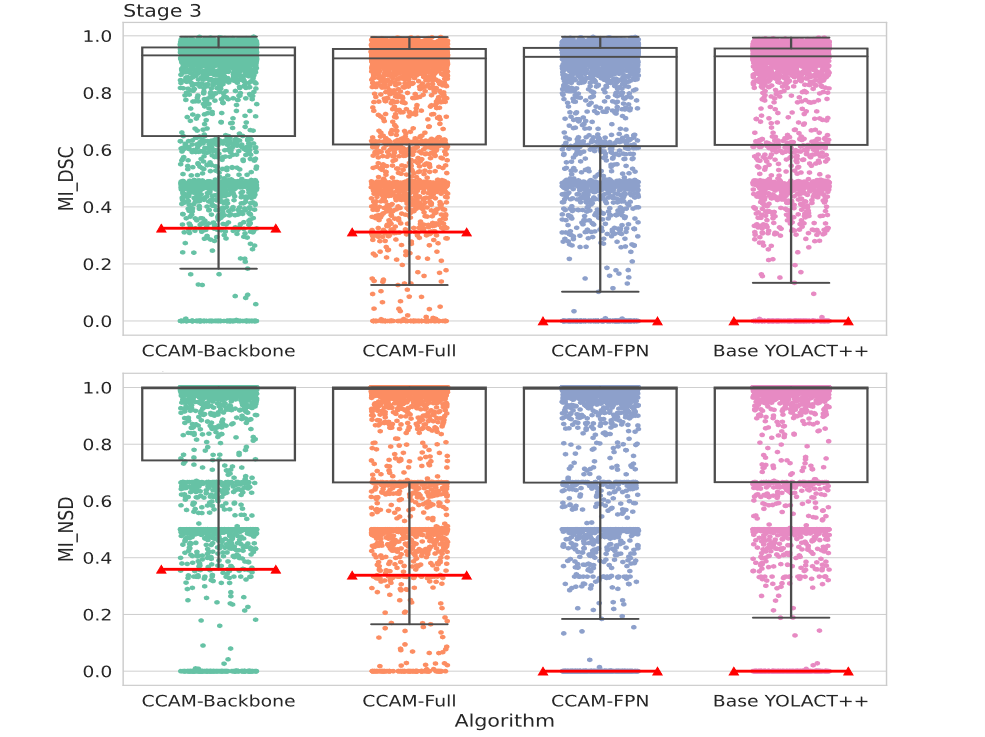}
    \caption{Dot-and-boxplots showing the individual performance of algorithms in stage 3 of the challenge. Red lines indicate the aggregated metric by 5\% percentile also reported in Table \ref{tab:metric_scores}.}
     \label{fig:boxplots}
\end{figure}

Figure \ref{fig:boxplots} shows dot-and-boxplots of the metric values for each algorithm over all test cases in stage 3 of the challenge. Among the three model variations to which we added attention modules, CCAM-Backbone achieved the best results in terms of robustness. This result indicates that the contextually enriched feature maps from the ResNet-101 backbone are powerful enough to generate more accurate mask prototypes and coefficients in the YOLACT architecture and ultimately better segmentation outputs. 

From our results, we can determine that adding attention mechanisms on the FPN outputs only increases the model performance slightly when compared to the baseline. However, this slight improvement becomes non-significant when considering the strict aggregated challenge metrics. 

Considering the good performance of CCAM-Backbone and the minor improvements of CCAM-FPN, one might believe that combining both configurations in CCAM-Full would lead to better results than having only one of them. However, as we can observe in their respective boxplots, this is not the case. An explanation of this behavior is that adding too many attention modules that integrate global context might lead to an over-mixing of information and noise, commonly known as over-smoothing, a common problem in graph neural networks from which CCAM takes inspiration.

\begin{table}
\centering
\caption{Aggregated evaluation performance for stage 3 of the challenge. The top section of the table includes the teams from the 2019 challenge. Team scores were taken from \cite{Ross_2020}. The bottom section includes our three attention models plus the baseline without attention.}
\label{tab:metric_scores}
\begin{tabular}{lccc}
\hline
\textbf{Team/Algorithm}                                                               & \textbf{MI\_DSC} & \textbf{MI\_NSD} & \textbf{FPS} \\ \hline
\textit{www} (Mask R-CNN)                                                             & 0.31             & 0.35             & 5*           \\
\textit{Uniandes} (Mask R-CNN)                                                        & 0.26             & 0.29             & 5*           \\
\textit{SQUASH} (Mask R-CNN)                                                          & 0.22             & 0.26             & 5*           \\
\begin{tabular}[c]{@{}l@{}}\textit{CASIA\_SRL} (Dense\\ Pyramid Network)\end{tabular} & 0.19             & 0.27             & 5*           \\
\textit{fisensee} (2D U-Net)                                                          & 0.17             & 0.16             & 18*          \\
\textit{caresyntax} (Mask R-CNN)                                                      & 0.00             & 0.00             & 5*           \\
\textit{VIE} (Mask R-CNN)                                                             & 0.00             & 0.00             & 5*           \\ \hline
CCAM-Backbone                                                                         & 0.313            & 0.338            & 49           \\
CCAM-Full                                                                             & 0.308            & 0.333            & 45           \\
CCAM-FPN                                                                              & 0.000            & 0.000            & 60           \\
Base YOLACT++                                                                         & 0.000            & 0.000            & 75           \\ \hline
\multicolumn{4}{l}{\begin{tabular}[c]{@{}l@{}}*Approximated from base method.\\ Original measurement was not reported.\end{tabular}}      
\end{tabular}
\end{table}


Next, we compare our proposed models and baseline to the top participants of the 2019 challenge (note that the 2020 edition did not take place). Table \ref{tab:metric_scores} shows the aggregated MI\_DSC and MI\_NSD values achieved for each participant/model, as well as the network architecture they were based on and their approximated/measured frame rate. Regarding the robustness of our method, CCAM-Backbone achieves competitive results in both metrics with respect to the top-performing contestant, reaching 0.313 MI\_DSC and 0.338 MI\_NSD compared to 0.31 and 0.35 respectively by team \emph{www}. 

It is important to note, though, that this team used data from the EndoVis 2017 challenge to improve their generalization performance on additional data. However, our best model outperforms the second-best contestant by a considerable margin: 0.053 MI\_DSC and 0.048 MI\_NSD.

An important contribution of our method is its ability to run in real-time. Inference speed performance was tested on a 10 second video snippet from the ROBUST-MIS dataset a total of ten times per model. The reported frame rates were then aggregated by the mean. Inference was tested on a single Tesla P100 GPU from the DGX-1 cluster with video multi-frame enabled.

As seen in Table \ref{tab:metric_scores}, the vast majority of the submitted models were based on Mask R-CNN, which is inevitably slow as it relies on a two-stage detector that limits its performance to 5 fps at most.  In contrast, our models comfortably fulfill real-time operation requirements for clinically usable models, running at $\geq45$ fps.  

\begin{figure}
   
    \includegraphics[width=.46\textwidth]{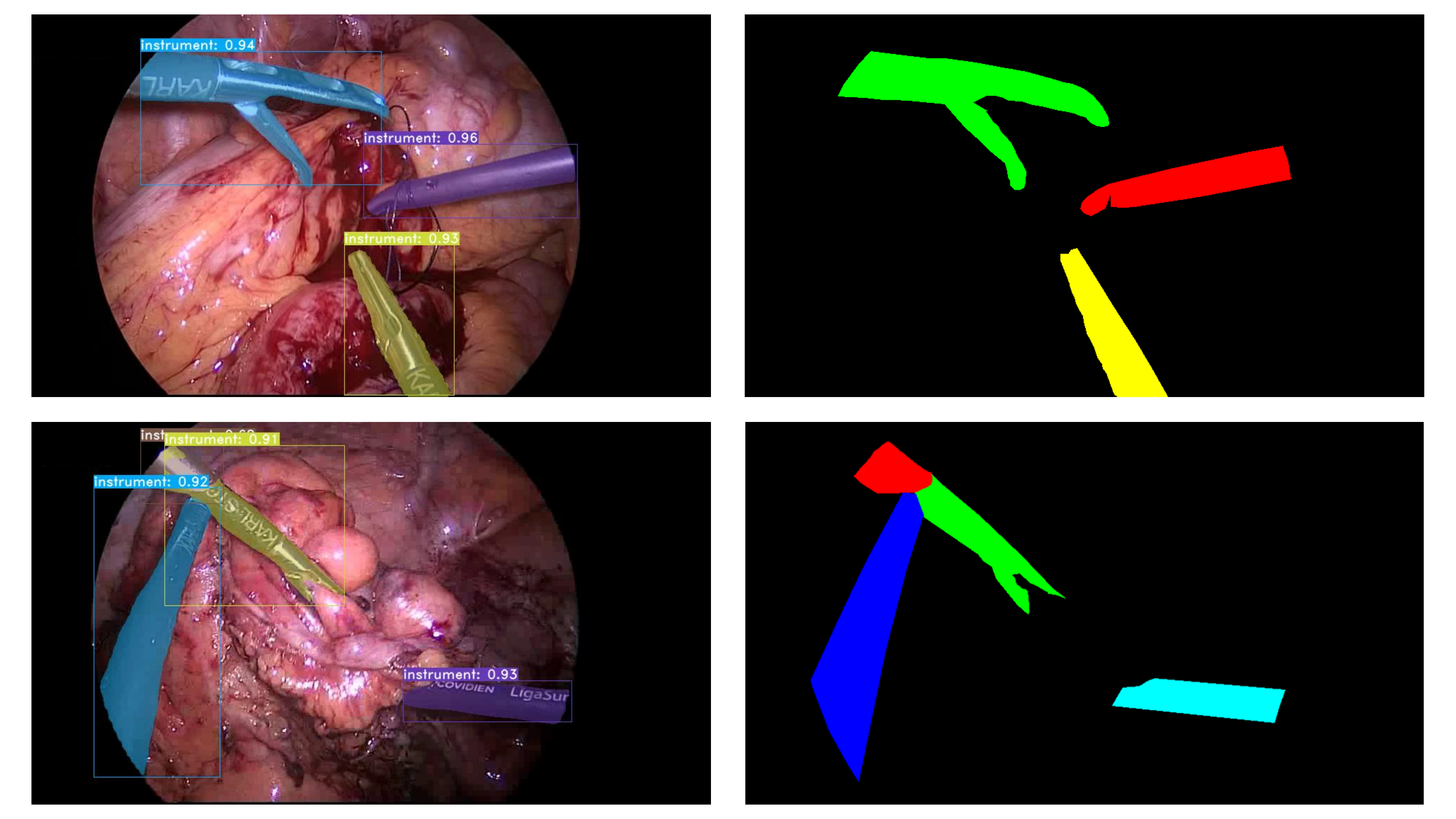}
    \caption{Left: CCAM-Backbone evaluation results on stage 3 frames. Right: Ground truth annotations. All images have the confidence threshold set to 0.3.}
     \label{fig:segmentations}
\end{figure}

\begin{figure}
    \includegraphics[width=.46\textwidth]{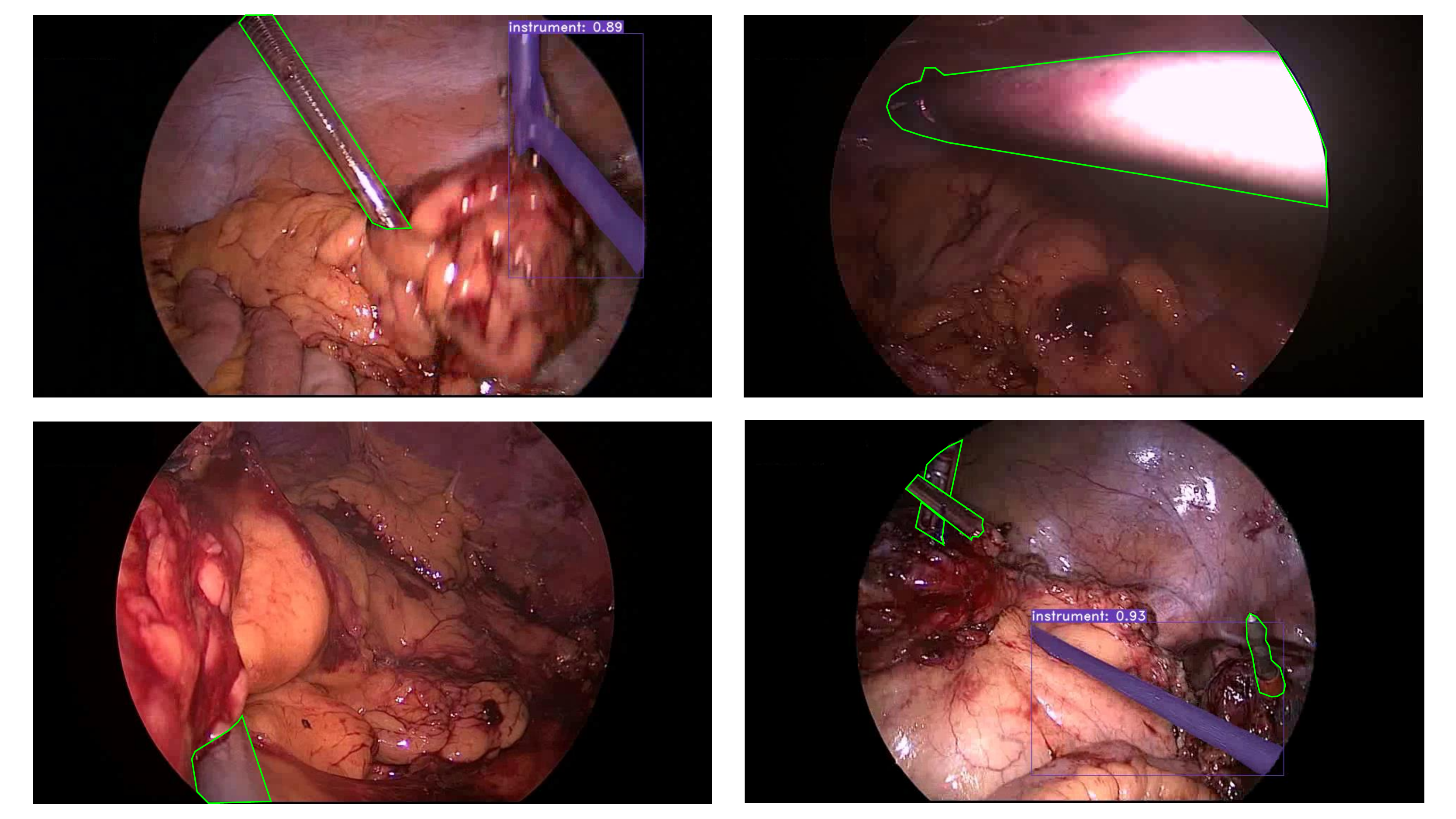}
    \caption{Examples of challenging frames, green contours indicate instrument instances not detected by our model. From left to right and top to bottom: 1. transparent instrument, 2. large reflection on instrument, 3. small instrument on the edge of the field of view, and 4. multiple instruments, some partially occluded and on the edge of the field of view.}
    \label{fig:challenging_frames}
\end{figure}

Notably, our top model produces high-quality and temporally consistent masks. Figure \ref{fig:segmentations} shows some examples with varying types and number of instruments together with their respective ground truth annotations. The model is robust to occluded instruments and various harsh conditions, like blood, smoke, and poor lighting. Nevertheless, it struggles with transparent instruments and small instruments on the edge of the field of view. Figure \ref{fig:challenging_frames} illustrates some examples of challenging frames for our proposed algorithm, which we will seek to address in future work.

\section{Conclusion}
\label{sec:conclusion}
We presented a novel approach for multi-instance segmentation of medical instruments based on the YOLACT architecture extended with embedded criss-cross attention modules. The addition of attention made it possible to extract better global context and exploit salient features leading to improved segmentation results. Our best model yielded competitive results in terms of robustness compared to the state-of-the-art, reaching 0.313 on area-based metric MI\_DSC and 0.338 on distance-based metric MI\_NSD while attaining real-time performance. Our experiments showed that adding attention modules to  YOLACT boosts the performance of the model and increases robustness. However, since CCAM is based on graph neural networks, it can potentially create feature clusters that can cause over-smoothing of the learned features. Thus, embedding attention modules at each layer may hinder the performance resulting from the over-mixing of global information and noise.


In future work, we plan to experiment with different types of attention mechanisms besides CCAM. Moreover, to increase the robustness of future models, we believe that stronger data augmentation aimed towards challenging instances could improve the model performances.


\section*{Acknowledgments}
The authors wish to thank the AI Hub and the CIIOT at ITESM for their support for carrying the experiments reported in this paper on their NVIDIA's DGX computer.


\addtolength{\textheight}{-12cm}   


\end{document}